\documentclass{article}

\usepackage{microtype}
\usepackage{graphicx}
\usepackage{subfigure}
\usepackage{booktabs} 

\usepackage{hyperref}
\usepackage[export]{adjustbox}

\usepackage[accepted]{icml2023}

\usepackage{amsmath}
\usepackage{amssymb}
\usepackage{mathtools}
\usepackage{amsthm}

\usepackage[capitalize,noabbrev]{cleveref}

\usepackage{balance}
\usepackage{flafter}
\theoremstyle{plain}

\theoremstyle{definition}

\theoremstyle{remark}

\usepackage[textsize=tiny]{todonotes}
\usepackage[acronym]{glossaries}
\glsdisablehyper
\newacronym{dnn}{DNN}{Deep Neural Networks}
\newacronym{mstde}{MSTDE}{Mean Squared TD Error}
\newacronym{rl}{RL}{Reinforcement Learning}
\newacronym{gvfs}{GVFs}{General Value Functions}
\newacronym{td}{TD}{Temporal Difference}
\icmltitlerunning{Discovering Object-Centric Generalized Value Functions From Pixels}

\begin{document}

\twocolumn[
\icmltitle{Discovering Object-Centric Generalized Value Functions From Pixels}



\icmlsetsymbol{equal}{*}
\icmlsetsymbol{sym}{\dag}

\begin{icmlauthorlist}
\icmlauthor{Somjit Nath}{ets,mila}
\icmlauthor{Gopeshh Raaj Subbaraj}{mila,udem}
\icmlauthor{Khimya Khetarpal}{sym,mila,mcgill}
\icmlauthor{Samira Ebrahimi Kahou}{ets,mila,cifar}
\end{icmlauthorlist}

\icmlaffiliation{ets}{École de technologie supérieure}
\icmlaffiliation{mila}{Mila-Quebec AI Institute}
\icmlaffiliation{udem}{Université de Montréal}
\icmlaffiliation{mcgill}{McGill University}
\icmlaffiliation{cifar}{CIFAR AI Chair \dag now at DeepMind}

\icmlcorrespondingauthor{Somjit Nath}{somjit.nath.1@ens.etsmtl.ca}

\icmlkeywords{Machine Learning, ICML}

\vskip 0.3in
]



\printAffiliationsAndNotice{}  



\begin{abstract}

Deep Reinforcement Learning has shown significant progress in extracting useful representations from high-dimensional inputs albeit using hand-crafted auxiliary tasks and pseudo rewards. Automatically learning such representations in an object-centric manner geared towards control and fast adaptation remains an open research problem. In this paper, we introduce a method that tries to discover meaningful features from objects, translating them to temporally coherent `question' functions and leveraging the subsequent learned general value functions for control. We compare our approach with state-of-the-art techniques alongside other ablations and show competitive performance in both stationary and non-stationary settings. Finally, we also investigate the discovered general value functions and through qualitative analysis show that the learned representations are not only interpretable but also, centered around objects that are invariant to changes across tasks facilitating fast adaptation.
\end{abstract}

\section{Introduction}\label{sec:intro}
Learning control from high-dimensional input such as images is a complex problem relevant to many real world applications. While researchers have made huge strides in Deep Reinforcement Learning (RL), decision making from images remains a challenge due to the difficulty of discovering meaningful features that are invariant across tasks. ~\citet{levine2016end,kalashnikov2018scalable} demonstrate how agents can learn a policy from pixels to be used in real-life applications. One of the standard practices in RL is to learn a control policy from pixels in an end-to-end fashion. The end-to-end learning paradigm presents certain downsides due to the black-box nature of \gls{dnn}. A significant challenge faced during learning from pixels is the inability to learn a good control policy from high dimensional visual inputs and limited information in a single scalar reward signal. To address this issue, recent works have shown that the agent performance can be improved by using auxiliary tasks alongside the main RL objective. The corresponding auxiliary losses can be trained using different objectives from supervised~\cite{jaderberg2016reinforcement,schwarzer2020data, guo2020bootstrap}, self-supervised~\cite{ oord2018representation,laskin2020curl}, or RL domains~\cite{veeriah2019discovery}. 

Designing meaningful auxiliary tasks poses many challenges; most previous works considered them in order to learn visual features entirely without rewards. For example, \citet{schwarzer2021dataefficient} learns visual features entirely without considering rewards.
In contrast, our focus here is to \emph{discover} auxiliary tasks that are \textit{reward-driven} and help learn useful representations which in turn can facilitate learning better policies. It is generally easier to define predictive knowledge of the environment through value functions, thus learning auxiliary tasks that are driven by such learned reward functions can help to learn many, different, potentially invariant, properties of the environment in order to be robust to any non-stationarity. Thus, these reward-driven auxiliary objectives could specifically serve an important purpose to learn and retain representations when learning in presence of non-stationarity. 
For instance, in Atari, fundamental objects and corresponding characteristics remain consistent when an agent transitions to a new unseen level.
Discovering these objects (features) and learning associated properties (cumulants) could drive generalization and adaptation in realistic continual learning or multi-task settings. 

This idea finds its origins in the Horde architecture~\cite{horde} where auxiliary objective constitutes learning value functions for a multitude of pseudo-rewards termed as \gls{gvfs}. GVFs provide a mechanism to learn value functions by choosing any scalar reward signal as a cumulant. These value functions can be used to extract useful predictions based on interactions with the environment thereby learning a rich representation. These estimates of environmental knowledge could potentially act as useful representations for an RL agent in different settings. In this work, we tackle the problem of learning General Value Functions in the pixel space, in an end-to-end manner by learning \textit{useful} representations. We design a framework for pixel-based agents to take advantage of GVF predictions by treating them as features to ensure a compact yet rich representation. Our main contributions are as follows:
\vspace*{-2mm}
\begin{itemize}
\setlength{\itemsep}{0 mm}
    \item We propose OC-GVFs: an end-to-end approach to automatically discover object-centric General Value Functions from pixels. Our method can subsequently be used as features for learning the downstream control policy. (Sec~\ref{sec:alg})
    \item Instead of learning a huge number of auxiliary tasks as GVFs and only discovering few relevant ones, we aim for our method to focus on discovering the key attributes of the environment. To do this, we consider the \textbf{limited GVF regime}, i.e. a small number of GVFs to ensure we get an information-rich representation of the environment.
    \item We empirically demonstrate that OC-GVFs can outperform the current state-of-the-art algorithms for GVF discovery in both stationary and non-stationary environments. (Sec.~\ref{sec:main_res})
    \item We show that the proposed method can quickly adapt to new unseen situations with increasing complexity. (Sec.~\ref{sec:transfer})
\end{itemize}

\section{Background \& Related Work}\label{sec:background}
\begin{figure*}[!ht]
    \centering
    \includegraphics[width=0.9\linewidth]{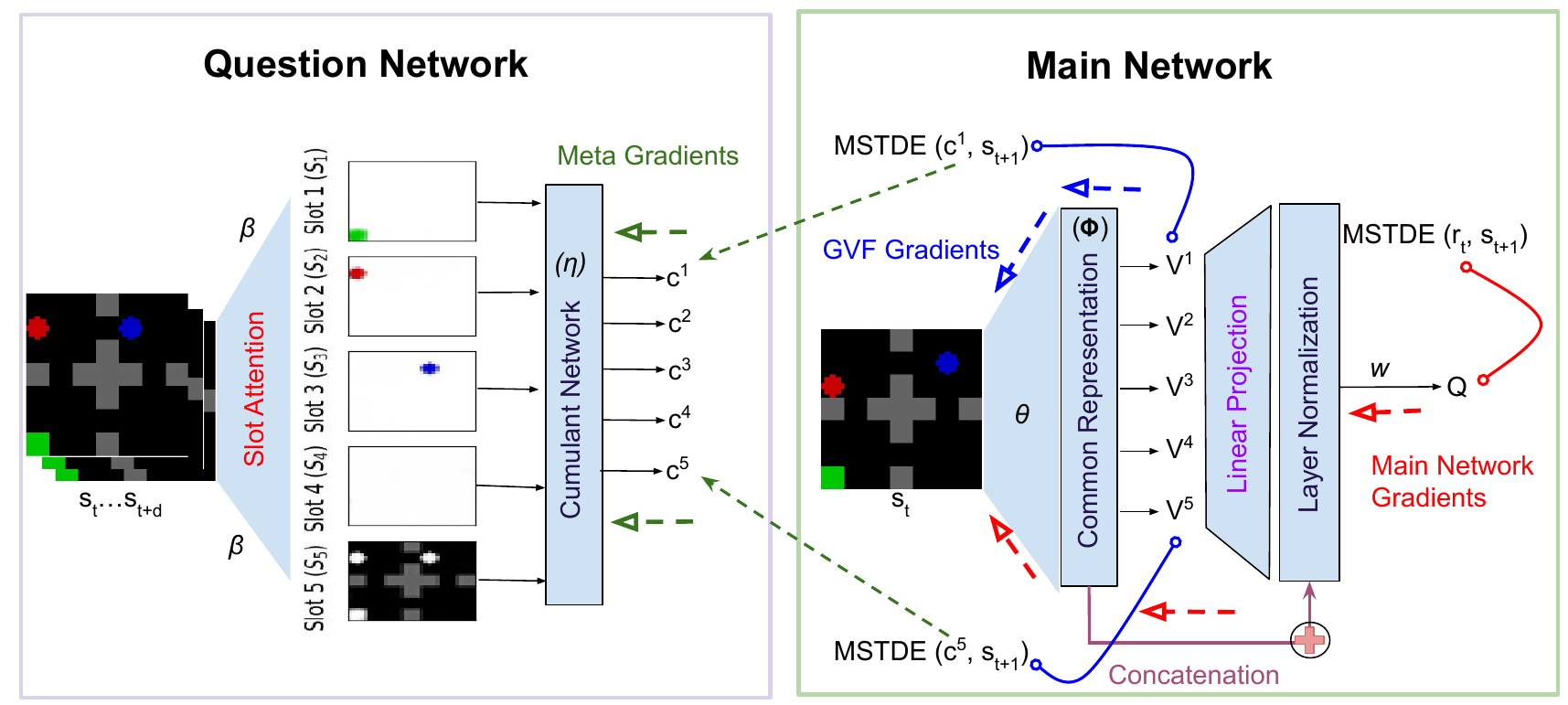}
    \caption{\textbf{The Object-Centric GVF Learning Framework}: We have two parts a \textit{Question Network} and a \textit{Main Network}. The Question Network takes in a batch of inputs ($s_t \dots s_{t+d}$) and tries to predict GVF questions (cumulants) from each slot. Each of the cumulants corresponds to the slot outputs ($c^{k}$). The Main Network is an RL algorithm with GVF heads each of which is trained by the cumulants from the Question Network. For training GVFs, we use the Mean Squared TD Error (MSTDE) which depends on the cumulants discovered by the Question Network ($\gamma$ is the same as defined for the main task). The outputs of the GVFs are projected into the latent space where it is concatenated with the representation before normalization and action value prediction with respect to the main task.}
    \label{fig:network}
\end{figure*}
\subsection{General Value Functions}\label{sec:gvf}
GVFs are value functions that are suited to represent predictive knowledge of an agent's environment, such as ``how far a particular object is in this gridworld'' which represent knowledge about the environment. 
In ~\citet{horde}, in addition to the main task, the Horde architecture considers many sub-agents called demons which learn different predictive components of the environment. 

GVFs are essentially same as the value functions defined in a Markov Decision Process, except the rewards are not obtained from the task specifically. The input to the GVFs would thus be a policy, a discount factor, and scalar rewards known as cumulants. These cumulants can be described as \textit{questions} to the GVFs. GVF \textit{answers} then are formalized as value functions, henceforth referred to as GVFs. In summary, GVFs are defined as the \textbf{expected discounted return over a certain trajectory, where the returns are defined as the discounted sum of the cumulants of interest.}
Analogous to value functions in RL, GVFs can be learned by any value-based RL algorithm e.g. Temporal Difference (TD) Learning. In this paper, we use the mean squared TD error, (MSTDE) ~\cite{rlbook}.

\subsection{Auxiliary tasks in RL}\label{sec:aux_task}
The idea of auxiliary tasks in RL was introduced to learn from signals other than just scalar reward signals especially when the rewards are sparse, delayed, or noisy. Auxiliary tasks are learned in parallel to the main RL loss~\cite{shelhamer2016loss, sutton2018reinforcement}.

Auxiliary tasks are an umbrella term and could refer to any task that can aid an RL Agent by predicting observations from the environment. Some of the common auxiliary task setups include reward prediction~\cite{Jaderberg17pbt}. In environments with sparse reward structures, auxiliary tasks provide instantaneous targets for shaping the representation in the absence of reward. Few other works focus on state prediction in the latent space and use the loss from this prediction to drive certain aspects like exploration in RL~\cite{pathak2017curiosity}. \citet{paster2021planning} focuses on modeling the inverse dynamics of the environment i.e. prediction of actions from state and next-state representations. \citet{veeriah2019discovery} presented a principled meta-gradient algorithm for the discovery of GVF-based questions to use as auxiliary tasks in the context of RL. 

\subsection{Cumulant Discovery}\label{sec:discovery}
To learn GVFs, we need a cumulant for defining the TD target. \textit{Discovery} refers to automatically learning cumulants, that aid in the primary task, referred as \textit{useful cumulants}.
Recall that (Sec.~\ref{sec:aux_task}),\citet{veeriah2019discovery} develop a framework to discover GVF \textit{questions} with a \textit{question network} and learn the cumulants (with the main network). This is in contrast to other auxiliary task methods in RL which use hand-crafted cumulants for learning in the environment~\cite{jaderberg2016reinforcement}. 
They propose a meta-learning approach to discover important questions that are useful for learning the main task in the environment and then estimate answers through GVFs for these discovered questions. The idea here is to use the same RL loss driven by the reward from the environment for the discovery of auxiliary tasks. The core intuition behind cumulant discovery is that through the availability of a question network with meta-learnable parameters, we enable the agent to discover useful questions directly from experience. We believe using the value functions learned from these discovered cumulants as part of the input representation for the main RL agent is crucial to our method. 

More recently, ~\citet{metagvf} augment the agent's observations with GVFs for control in RL. \citeauthor{metagvf} also integrate the discovery of GVFs and their use in a single end-to-end framework using a meta gradient descent approach. In a nutshell, they shape the GVF predictions based on the control agent’s learning process and use those predictions directly as features for learning a better control policy. In contrast, we only focus on cumulant learning, but we can easily extend our framework to learning $\gamma$ and policy correction parameters~\cite{metagvf}.

\subsection{Object-Centric Representations}
There have been several recent works that tackle challenging tasks such as object manipulation in presence of multiple objects~\cite{watters2019cobra, van2019perspective, veerapaneni2020entity}. Much of this work is in a single-task RL setting for a particular reward signal.
In contrast, certain works have also focused on learning object-centric representations from images in multi-task RL settings~\cite{pong2019skew, nair2018visual, ghosh2019learning, warde2018unsupervised}. 
These methods rely on some assumptions that the observations can be encoded into a single vector which makes it harder to learn in environments with multiple objects. \citeauthor{zadaianchuk2020self} proposed learning object-centric representations which are used for reward shaping. They claim that this approach leads to solving tasks independently and then combining these skills during evaluation. 

We chose the slot attention mechanism~\cite{slotattn} to learn object centric representations in our approach. These slot representations are used to learn cumulants (through the cumulant network, See Sec.2.3) which in turn are used as part of the input representations to learn a policy. Moreover, there is a one-to-one mapping between the slot representations and the cumulants. This enables discovering useful questions from all the captured slots and ensures that no part of the original image is left unattended.

In this paper, we aim to ask the question: \textit{What are these discovered features learned by these networks and can they be used as input features?}  \citeauthor{metagvf} experimented with a similar architecture for non-image based domains where they concatenated the states with learned GVFs. In our work, we extend this methodology to more realistic image-based domains which are prevalent in RL. A big part of question discovery in these papers is driven by the main RL loss, however, RL loss on its own cannot capture object-centric representation efficiently. Since RL environments heavily rely on object semantics, we add another object discovery loss to the discovery network to bind the discovered cumulants to certain objects discovered by this architecture as explained in Sec.~\ref{sec:slots}.

\section{Discovering Object-Centric GVFs}\label{sec:alg}
\begin{algorithm}[!ht]
\begin{minipage}{\linewidth}
    \begin{algorithmic}
    \STATE \textbf{Input parameters}
    \STATE Num of Slots $N$,  Num of Training Steps(slot module) $M$
    \STATE Observations $O$, Num of GVFs $K$, Num of episodes $E$
    \STATE Initialize parameters of networks $\beta$, $\theta$, $\eta$
    \STATE Initialize learning rate of networks $\alpha_{1}$, $\alpha_{2}$, $\alpha_{3}$
    \FUNCTION{Slot Module Training Phase ($N$, $O$, $\beta$, $M$, $\alpha_{1}$)}
        \FOR {$i  \gets 1$ to $M$}
            \STATE $S_{i} \gets$ slot\_model$(O_{i})$
            \STATE $\hat{O_{i}}  \gets$ reconstruct$(S_{i})$
            \STATE $\beta_{t+1} \leftarrow \beta_t-\alpha_{1} \nabla_\beta\mathcal{L}\left({O_{i}},\hat{O_{i}}\right)$
        \ENDFOR
    \ENDFUNCTION
    \FUNCTION{GVFs Training Phase($K$, $E$, $\theta$, $\eta$)}
            \FOR {$n  \gets 1$ to $E$}
                \STATE ${t} \gets 0$
                \STATE ${done} \gets$ False
                \STATE $\theta_{n,0}\leftarrow \theta_{n}$
                \WHILE{not $done$}
                    \STATE ${t} \gets {t+1}$
                    \STATE $\{C_n^1,\dots,C_n^k\} \gets f_1(\eta; \{O_{t},\dots,O_{t+d}\})$
                    \STATE $\{\phi_t\} \gets f_2(\theta; O_{t})$
                    \STATE $\{V_1,\dots,V_k\} \gets f_3(\theta; \phi_t;\{C_n^1,\dots,C_n^k\})$
                    \STATE $\{\psi_1,\dots,\psi_k\} \gets$ proj$(V_1,\dots,V_k)$
                    \STATE $\chi_t \gets$  concat$(\{\psi_1,\dots,\psi_k\}; \phi_t)$
                    \STATE ${Q_t} \gets f_4(w, \chi_t)$
                    \STATE $\theta_{n,t} \leftarrow \theta_{n,t-1}-\alpha_{2} \nabla_{\theta_{n,t-1}}\mathcal{L^{RL}}\left(\theta_{n,t-1}\right)$
                \ENDWHILE
                \STATE $\eta_{n+1} \leftarrow \eta_n-\alpha_{3} \nabla_\eta\sum_{j=1}^{t}\mathcal{L^{RL}}\left({\theta_{n,j}}\right)$
                \STATE $\theta_{n+1} \leftarrow \theta_{n,t}$
            \ENDFOR
    \ENDFUNCTION
    \end{algorithmic}
    \end{minipage}
\caption{Object-Centric GVFs (OC-GVFs)}
\label{alg:Algorithm 1}
\end{algorithm}

Our proposed architecture consists of two separate networks: the \textit{question network} and the \textit{main network}. This two-network meta-gradient approach was introduced for GVF discovery by~\citet{veeriah2019discovery}. The primary difference in our work is that we use an Embedded Self prediction (ESP)~\cite{lin2021contrastive} type model to embed GVFs as useful features in our training pipeline. In the original ESP paper, the core intuition was that, if human understandable features are given to the model, the corresponding GVFs would capture meaningful properties of the policy. In a similar way, we directly adapt these trained GVFs as features of the agent's main value function.

We introduce two key modifications to this architecture for adaptation to all possible input spaces and better stability, namely 1) a key design decision in our approach is to concatenate GVFs with the states in the latent space. This is achieved via linear projection from the outputs of each of the GVFs into the latent space after which they are concatenated with the common representation from the main task. This modification removes the necessity for the states to be vectors which was the case for concatenation with directly the state inputs~\cite{metagvf}, and 2) addition of layer normalization~\cite{layernorm} after the concatenation. This helps especially in stability when the slots and hence the cumulants are not learned yet during the early phase of training. Next, we describe the individual components in detail.

\subsection{Question Network}\label{sec:qn}
The question network (Fig.~\ref{fig:network} Left) takes in a batch of state observations ($s_t \dots s_{t+d}$) which are unrolled from the replay buffer as inputs. These inputs are fed to a slot attention mechanism (parameterized by  $\beta$) that outputs slots, $S_i$ corresponding to discovered objects from the images. Each of these slots can be considered to have some features of the objects in the images. We learn slot representations through forward propagation of the question network. These slot representations are then mapped to each GVF cumulant using an MLP (parameterized by the meta-parameters $\eta$). The slots are trained by reconstruction loss as in~\citet{slotattn} and the cumulants are trained with the main RL loss similar to~\citet{veeriah2019discovery}. This forces the cumulants to capture task-specific properties of each object discovered by each slot. As discussed in the previous section, a GVF question is specified by a cumulant function, a discount function, and a policy. In our method, the question network only explicitly parameterizes the cumulants. Though this is a departure from the architecture used in~\citet{veeriah2019discovery} which parameterized both the cumulants and their corresponding $\gamma$, our method similar to~\citet{metagvf} parameterizes only the cumulants. This is because we wanted to capture the properties of objects which would be relevant to long-horizon settings, but this is definitely something that can be incorporated very easily into this architecture. For our experiments, we use the same $\gamma$ of the main agent for learning the GVFs.

\subsection{Main Network}\label{sec:main}
The main network (on the right side in Fig ~\ref{fig:network}), deals with the training of the GVF answers (parameterized by $\theta$) i.e computing the appropriate value function for each of the cumulants generated by the question network and also learning the main action-value function of the main task (parameterized by $w$). Note that the number of GVFs is a hyper-parameter in this setup. Once these losses are computed, we do the first backpropagation (blue line) update on the parameter theta as in regular gradient descent. An important detail here is that we do another backpropagation (red line) update on theta while updating the main agent based on the MSTDE with respect to the main task. The meta-parameters are updated based on the cumulative loss incurred over the unrolled data from the buffer. The number of steps of unrolled data is again a hyper-parameter that is tuned as part of the experiments.

\textbf{Action and State Value GVFs}: Before proceeding further, it is imperative to discuss the utility of both using action value GVFs and state value GVFs. Action value GVFs generally contain more information regarding each action than an expectation over actions in the case of state values. However, for our architecture, we preferred to use state value GVFs as they are less prone to divergence from off-policy Q-learning updates which can often be the case in~\citet{veeriah2019discovery}. However, since the authors of that paper only use them for learning the representation it does not affect them as much. More details including some empirical performance plots are in Appendix~\ref{sec:apndx_vq}.

To summarize, our method uses feature representations that are output from the slot attention module rather than using human-designed features. We believe this is a key component as we would like the GVFs to automatically discover useful characteristics in an environment. Since many RL environments are image-based and certain aspects of the environment do not change (like objects present in the environment), we believe a mechanism like slot attention adds the most value here. The slot attention module takes in representations from convolution layers and produces abstract representations called slots which bind well to objects in the visual inputs. We refer the reader to the original slot attention paper for more details ~\cite{slotattn}. However, since we have access to an experience replay buffer which is generally present in most RL algorithms, we do not require any pre-training for the slot attention. The data to train slot attention can be directly obtained from the buffer and thus we obtain an end-to-end training pipeline.

\section{Experiments}\label{sec:exp_details}
Next, we discuss the empirical performance of our algorithm across different domains and settings. We address the following questions: \textbf{Q1.} How does our approach compare to simple baselines in stationary environments? Can this approach significantly outperform other baselines in non-stationary tasks? (Sec.~\ref{sec:main_res})  \textbf{Q2.} Can learned object-centric representations adapt quickly to unseen tasks? (Sec.~\ref{sec:transfer}) \textbf{Q3.} How much do object-centric representations help in cumulant learning when compared to other architectures with similar feature discovery? (Sec.~\ref{sec:esp}). To address these questions, we first describe the domains and settings. In Sec.~\ref{sec:slots}, we explain our choice of using slot attention for object discovery. In Sec.~\ref{sec:norm},~\ref{sec:feats} we discuss the importance of layer normalization and leveraging the learned GVFs as features respectively.

\subsection{Domains}\label{sec:env} Visualization of the domains are in Appendix~\ref{sec:apndx_domains}.

\textbf{Collect-objects Environment:} is a customized version of the four-room gridworld environment similar to the one used in~\citeauthor{veeriah2019discovery}. The agent is rewarded for collecting objects  of different colors in the right order. The agent moves deterministically in one of four possible directions. For each episode, the starting position of the agent is chosen at the bottom left. The agent receives a reward of +5 for picking up the red objects and a reward of +10 for picking up the blue object after the red one. We also explore a non-stationary version of this domain where the locations of the objects spawn randomly inside their respective rooms after every episode. If the agent picks up the green object before the red one, the agent does not receive any reward.


\textbf{MiniGrid-Dynamic Obstacles:} For the experiments on non-stationarity, we used the MiniGrid Dynamic Obstacles~\cite{gym_minigrid}. In this domain, the agent is placed in a grid where it has to avoid colliding with obstacles and reach the goal. The starting position of the agent and the obstacle positions are all chosen at random.

\textbf{CoinRun \& StarPilot:} are a part of procedurally generated environments called ProcGen~\cite{procgen}. In CoinRun, the agent is tasked to capture the coin while avoiding obstacles. In StarPilot, the agent must destroy enemies while avoiding enemy fire and obstacles. We studied the performance of individual levels in CoinRun. In our task adaptation experiments, we study performance at a new unseen level after every episode in a sequential manner. 

\subsection{Settings}\label{sec:settings}
\begin{itemize}
\setlength{\itemsep}{0 ex}
    \item \textbf{Learning in the absence of non-stationarity.} The agent has to collect two objects in the gridworld. The positions of the two objects are fixed across episodes and unchanged. The agent needs to collect the two objects in the same order to receive full rewards. 
    \item \textbf{Learning in the presence of non-stationarity.} For Collect Objects, the objects spawn randomly. This creates a non-stationarity in the reward function across these tasks. The task boundaries are not known to the agent as well and are randomly changed after a fixed number of episodes.
    \item \textbf{Quick Adaptation.} To evaluate the agent's ability to quickly adapt to novel situations, we introduce the agent to new unseen levels in the CoinRun and StarPilot domains over time without any prior task information.
    \vspace*{-3mm}
\end{itemize}

\begin{figure*}[!ht]
\centering
\includegraphics[width=0.9\linewidth]{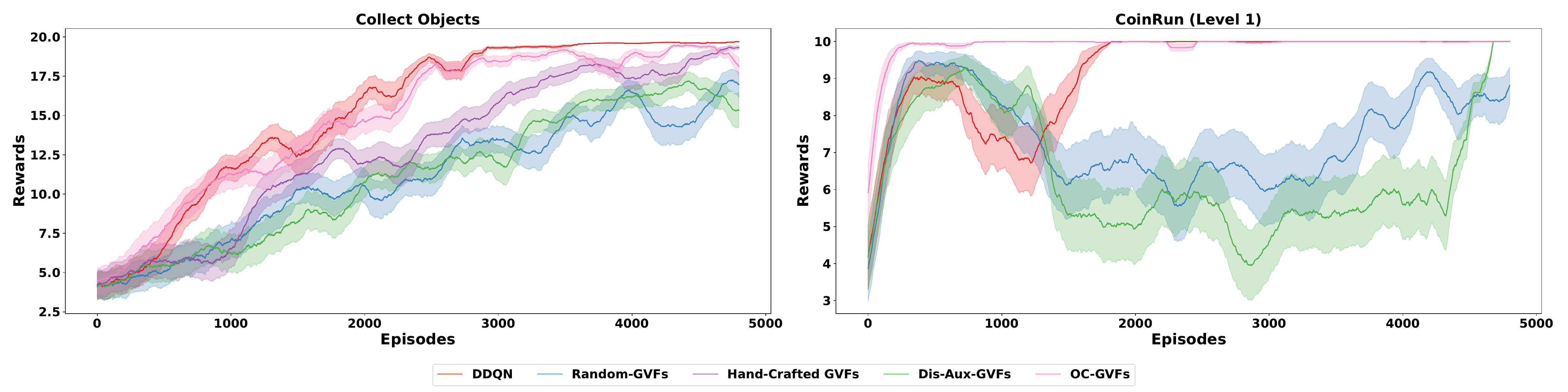}
\caption{\textbf{Learning in the absence of non-stationarity} shows that our method (OC-GVFs) is more sample efficient than using Random-GVFs and Dis-Aux-GVFs. All baselines are expected to show similar performance due to the simple nature of both the Collect Objects and CoinRun stationary domain here. In a simple task, DDQN is marginally better than other methods. However, in CoinRun which is more challenging OC-GVFs is significantly better than all other approaches. Shaded regions correspond to the standard error across $10$ independent runs.}
\label{fig:stat}
\end{figure*}
\begin{figure*}[!ht]
\centering
\includegraphics[width=0.9\linewidth]{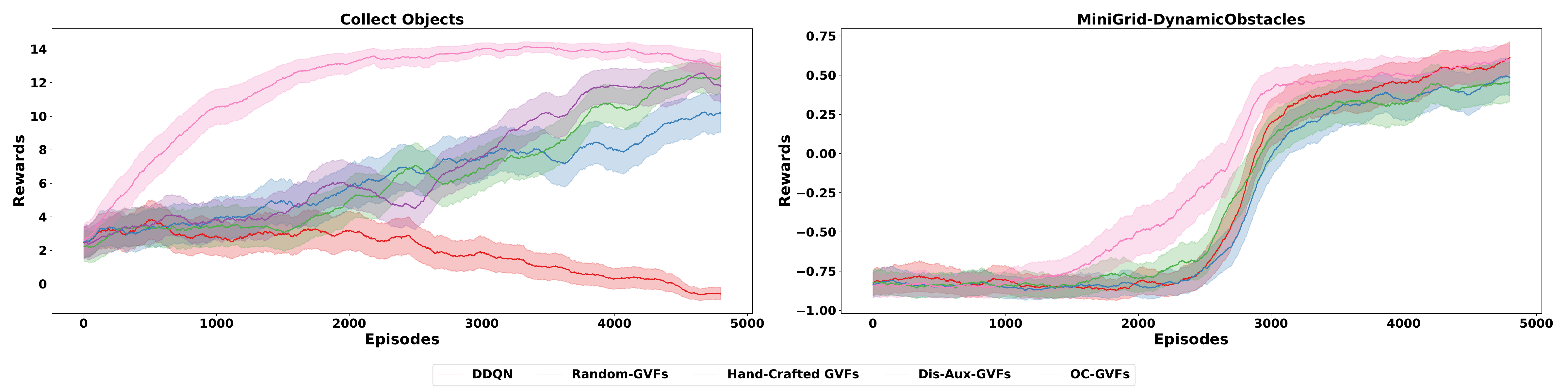}
\caption{\textbf{Learning in the presence of non-stationarity} shows that our approach is sample efficient and quick to adapt when the goal locations are dynamically changed after every episode in the CollectObjects environment. We also observe a similar trend in the Minigrid Dynamic Obstacles environment where the obstacle locations changes across episodes. Baselines such as DDQN cannot adapt to the non-stationarity induced due to changing goals as quickly as OC-GVFs across both domains. All results reported over 10 seeds with shaded region showing the standard error.}
\label{fig:non-stat}
\end{figure*}
\vspace*{-3mm}
\subsection{Baselines} \label{sec:baselines}
We consider the following baselines: 1) \textbf{DDQN}, which serves as the main RL algorithm, 2) \textbf{Random-GVFs} which uses randomly initialized Question Network, 3) \textbf{Hand-Crafted GVFs} that uses human-defined cumulants based on the task,\footnote{For Collect Objects, one of the cumulants essentially specifies the location of the red goals.} and 4) \textbf{Dis-Aux-GVFs}~\citep{veeriah2019discovery}, which is the only\footnote{To the best of our knowledge.} prior work that integrates the discovery of cumulants in the pixel space. A key motivation for our method is to ensure that the approach for discovery does not require a huge amount of data, in lieu of which we compare to other baselines in the \textbf{limited GVF regime}. We limit the number of GVFs to be $5$ for all the environments, the same as the number of slots. We include all the relevant hyper-parameters and implementation details in Appendix~\ref{sec:apndx_implementation} and open-source the code\footnote{{\url{https://github.com/Somjit77/oc_gvfs}}}. Additionally, details of the baselines can be found in Appendix~\ref{sec:apndx_baselines} along with methods that involve GVFs as features.

\subsection{Our approach performs competitively in both stationary and non-stationary settings. (\textbf{Q1.})}\label{sec:main_res}
In this section, we show the performance of our algorithm (Sec.~\ref{sec:alg}) comparing to the baselines described in Sec.~\ref{sec:baselines}. We here demonstrate learning in the presence of stationary and non-stationary settings.

\textbf{Stationary Domains}: We note that for simple environments like Collect Objects and CoinRun (first level), DDQN performs very well because the task comprises of fixed locations of objects. As seen in Fig.~\ref{fig:stat} (a), for Collect Objects all the baselines perform similarly, with OC-GVF slightly better than all the other baselines with GVFs. In Fig.~\ref{fig:stat} (b), we notice the best performance for our algorithm with DDQN also converging within 2000 episodes. Since this task does not have a lot of \textit{auxiliary} tasks to discover, Dis-Aux-GVFs are not as fruitful in accelerating performance in the main task. 

\textbf{Non-Stationary Domains}: These experiments highlight the benefits of using object-centric representations as features. This is evident in Fig.~\ref{fig:non-stat} (a) where our algorithm (OC-GVFs) is able to converge much faster than all the other baselines. In the Dynamic Obstacles Environment (Fig.~\ref{fig:non-stat} (b)), the difference is not as noticeable as the task is relatively easier, however, our method still outperforms other baselines.

\subsection{OC-GVFs is amenable to fast adaptation in the presence of increasing complexity in tasks. (Q2.)} \label{sec:transfer}

\begin{figure*}[!ht]
\centering
\includegraphics[width=0.9\linewidth]{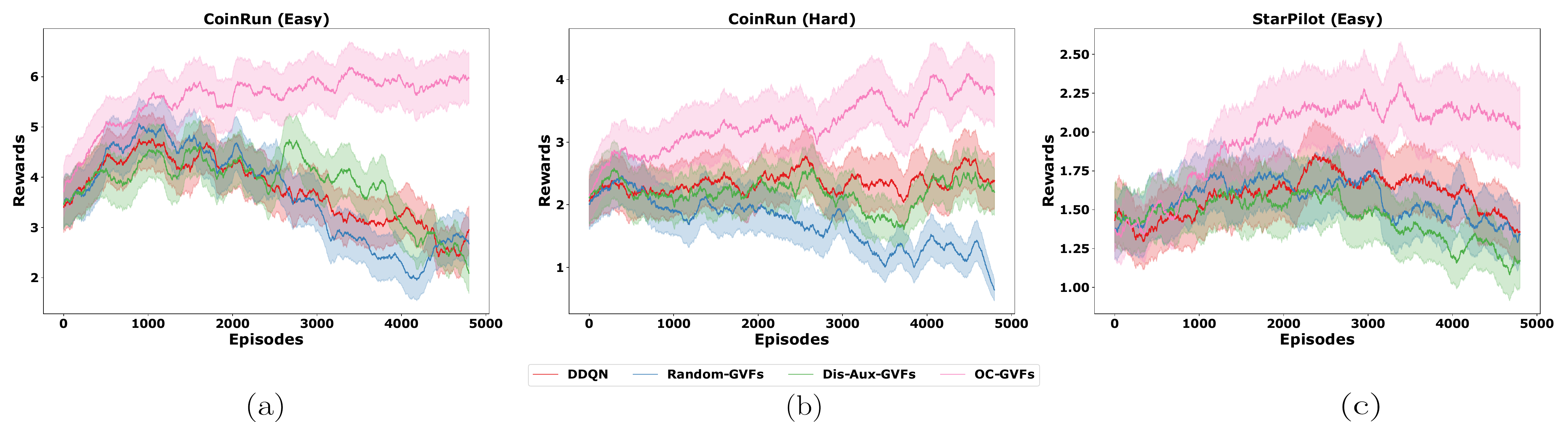}
\caption{\textbf{Adaptation to new tasks:} Our approach OC-GVFs can tackle changing levels better than the baselines. In these experiments, we sample a new task with different difficulty levels after every episode from the first 50 levels. This sampling is carried out from either an easy-to-learn distribution or a hard distribution which is slightly more challenging. The baselines cannot adapt as quickly and perform similarly to DDQN, suggesting no improvement in performance with GVFs. We compare with the baselines mentioned above and report results over 10 seeds.}
\label{fig:adaptation}
\end{figure*}
GVFs help in learning predictive knowledge about the environment. As such learned GVFs can often help in adapting to a new task because the agent can utilize previously learned information and quickly adapt to new scenarios. Most end-to-end GVF learning schemes involve discovering cumulants with the main RL loss~\cite{veeriah2019discovery}. While this is useful for the main task, the utility of such GVFs is lost during nonstationarity. On the contrary, as we decouple object discovery (via reconstruction loss) and cumulant learning (via main RL loss) in our approach, Object-Centric GVFs can adapt and generalize to new situations much faster. During adaptation to a new task, OC-GVFs need to re-learn the GVFs corresponding to the new slots, however, most of the pre-existing slots can be re-utilized when the task complexity increases and the difficulty changes.

Task adaptation refers to adapting to unseen tasks, namely with level changes once algorithms have more or less converged then performance is reported on a new level. These transfer learning experiments are in Appendix~\ref{sec:apndx_transfer}. We also test the methods for more challenging adaptation by presenting the agent with new unseen levels after \textit{every} episode. This is much more complex for all the algorithms, as seen from Fig.~\ref{fig:adaptation}, where the baselines do not perform as well including DDQN which was performing well for single levels and transfers across levels. Although OC-GVFs see performance drops due to adaptation to unseen levels every episode, it accumulates a higher average reward across all the sampled levels compared to the baselines.

\begin{figure}[!ht]
    \centering    \includegraphics[width=\linewidth]{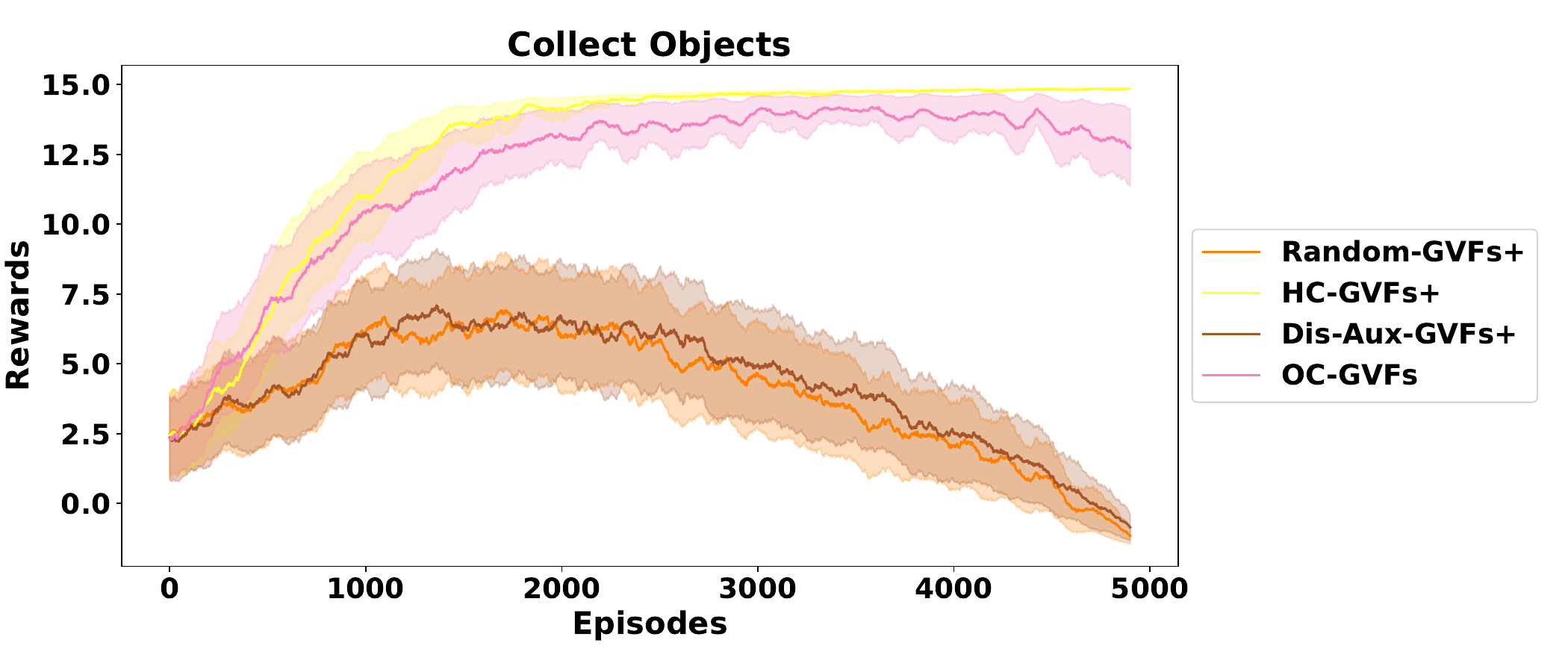}
    \vspace*{-3mm}
    \caption{\textbf{Object-centric representations} shows the comparison of our approach with other baselines while using GVFs as features($+$). This figure singles out the utility of object-centric representations from slot attention mechanism for discovery.}
    \label{fig:esp}
    \vspace*{-3mm}
\end{figure}
\subsection{Feature Discovery without object driven cumulant learning (\textbf{Q3.})}\label{sec:esp}
In this section, we aim to highlight the utility of discovering object-centric cumulants to be used as features. Discovering task-relevant cumulants has proven to be beneficial for learning representations. However, for using GVFs directly as features, it is essential that the cumulants discovered are tethered to some properties of the environment. We compare all the baseline algorithms to investigate whether these learned GVFs can be useful as features (described in Appendix.~\ref{sec:apndx_baselines}). Although GVFs discovered for representation learning in~\citet{veeriah2019discovery} were designed for auxiliary tasks, the discovered GVFs from this approach do not work well when used as features.Fig.~\ref{fig:esp} highlights the disparity between features learned with OC-GVFs and other baselines.  Hand Crafted GVFs can learn features best because it can exploit human information and results in best performance.

\begin{figure*}[!ht]
    \includegraphics[width=\linewidth, height=0.35\linewidth, left]{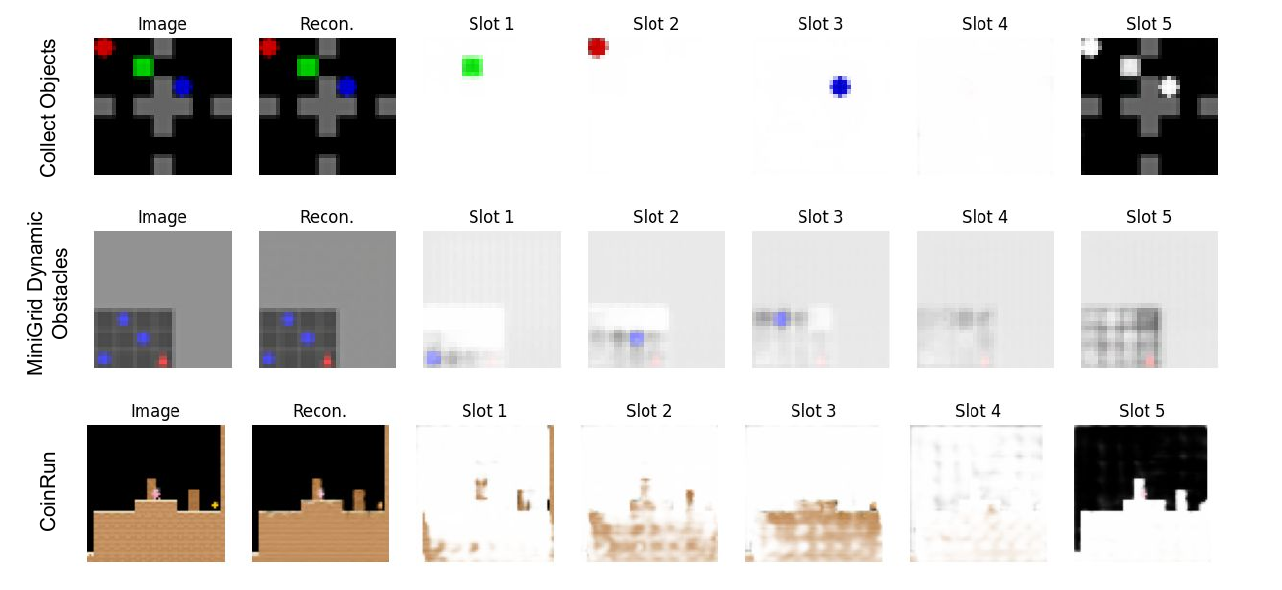}
    \vspace*{-7mm}
    \caption{\textbf{Slot Attention outputs} shows slots captured by the slot attention mechanism on states sampled from domains. The figure also shows learned state reconstructions for the sampled states across these environments.}
    \label{fig:slot_attention}
\end{figure*}

\begin{figure*}[!ht]
    \centering    
    \includegraphics[width=\linewidth]{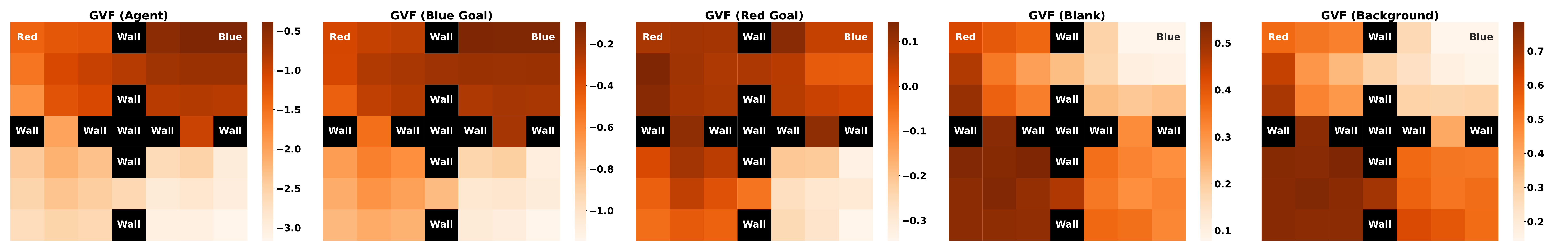}
    \caption{\textbf{Visualization of Learned GVFs:} We compare the GVFs learned by OC-GVFs in the CollectObjects environment trained with changing goal positions at every episode. Each of the GVFs is classified based on the slots that are fed as cumulants for training the respective GVFs. For example, the GVF that is learned by the cumulant defined by the slot that captures the Red Goal is GVF (Red Goal). The heatmaps are plotted based on the locations of the agent in the grid with the Red and Blue goal being constant throughout the evaluation. GVF(Blue Goal) and GVF (Red Goal) has higher values near the blue and red goal respectively which highlights it has learned some properties of those objects.}
    \label{fig:gvf_viz}
\end{figure*}

\section{Discussion}
\subsection{Qualitative Analysis: Understanding slots and how it translates to object properties}\label{sec:slots}

One of the crucial components of our approach is the discovery of objects with slot attention. Slot attention is generally trained with the reconstruction loss whereby each slot captures different objects present in the images. For most RL tasks, particularly from pixels, identifying objects can be a big overhead which we are delegating directly to a much more robust model. Since this is a vital component of our framework, it is imperative for slot attention to capture objects on which our cumulant learning is based. In environments where it is difficult to determine objects via slot attention, these methods will not work as well. This is one major limitation of the current framework, however, we have found slot attention to generalize well to different domains with proper architecture changes.

From Fig.~\ref{fig:slot_attention}, we observe how slot attention can segregate the main objects of the state by assigning each slot to those objects. Once features of these objects are learned, they can be used to learn cumulants pertaining to these objects. Slot attention can generalize easily to different locations of such objects which is what enables cumulant learning much faster in comparison to other baseline methods. Another important characteristic of slot attention is that we can be really flexible with the number of slots. In Fig.~\ref{fig:slot_attention}, for both the environments, there is an empty slot that does not contribute to capturing an object, but the relevant objects are still captured in the other slots. 

\subsection{Visualization of Learned GVFs} \label{sec:gvfviz}
Understanding what each GVF actually learns is a important and adds interpretability to our feature design which was missing from previous works like~\citet{veeriah2019discovery}. However, adding visualizing GVFs learned from images is tricky because they map images to a scalar value which is difficult to represent in a plot. However, for the Collect Objects domain, the location of the agent can find be a good proxy for the entire observation image provided the other locations remain constant. This is what we use to design our visualization in Figure~\ref{fig:gvf_viz}. Each square represents an image with the location of the agent and Red and Blue goal locations are mentioned by text. Now, we plot the GVFs learned in the CollectObjects domain with non-stationary rewards. Each of the GVFs is defined by its own cumulants, which in turn is dependent on the slots that the slot attention model captures. In Figure~\ref{fig:gvf_viz}, we add the objects that the slots capture for easy understanding. From the Figure, it is quite evident that for each corresponding object, the GVFs have higher values near the vicinity of that object. This insinuates that the GVFs have learned some form of a distance metric to these objects which can be thought of as compact and rich features when concatenated with the main feature representation. This makes the GVF learning much more interpretable and helps explain the impressive performance of OC-GVFs.

\subsection{Importance of Layer Normalization} \label{sec:norm}
\begin{figure}[!ht]
    \centering    \includegraphics[height=0.45\linewidth, width=0.8\linewidth]{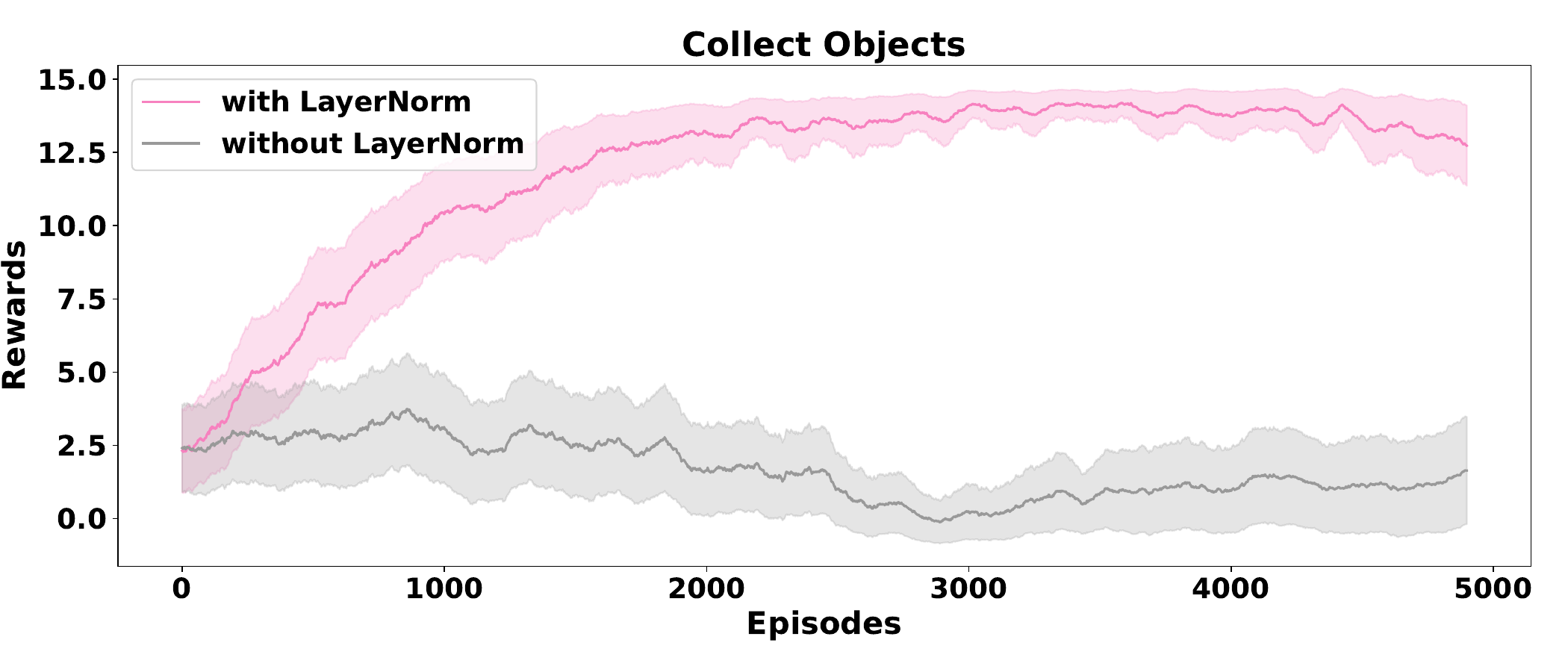}
    \caption{\textbf{Layer Normalization} shows how introducing layer normalization in the architecture creates a way for stable and faster learning. This is a critical factor that enables learning GVFs as features.}
    \label{fig:norm}
\end{figure}

One of the crucial elements of the proposed architecture is adding a layer normalization layer to the feature space. This is particularly important as in the early phase of learning the slots discovered can be poor leading to absurd cumulant discovery which would make the learned value functions erratic and unstable. This problem would not be as evident for algorithms that only use GVFs for learning representations as they do not directly affect the main value function. As affirmed by Fig.~\ref{fig:norm}, adding layer normalization can significantly improve the stability of our framework leading to faster learning.

\subsection{Utility of having GVFs as Features}\label{sec:feats}
\begin{figure}[!ht]
    \centering    \includegraphics[height=0.45\linewidth, width=0.8\linewidth]{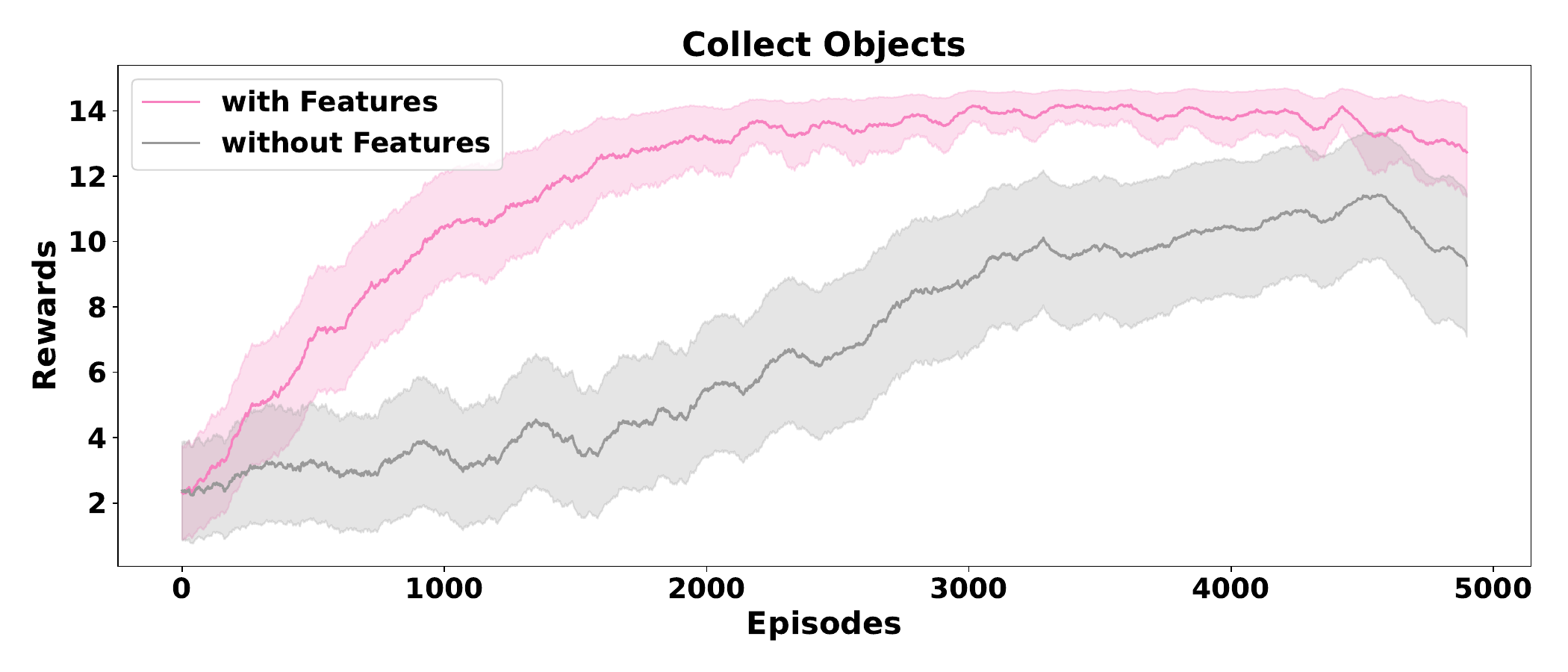}
    \caption{\textbf{Utility of GVFs as Features:} We compare the performance of an RL agent using GVFs for learning representations versus using GVFs as features. Even though both the GVFs capture object-centric representations, using the learned GVFs as features makes the most difference.}
    \label{fig:without_esp}
\end{figure}
Another important aspect of the proposed architecture is utilizing GVFs as a part of the feature space. This can be really helpful especially when the learned GVFs contain predictive knowledge about the environment which can be directly utilized by the agent.  When the cumulants are well-defined then this approach really shines. Fig.~\ref{fig:without_esp} demonstrates the performance of OC-GVFs with and without features. For OC-GVFs without features, we have slot attention capture object properties to be used for representation learning only.

\section{Conclusion and Future Work}
In this work, we showed the effectiveness of object-centric representation in discovery of GVFs that are used for the control in reinforcement learning.
Moreover, we also demonstrated how these learned GVFs capture important components in visual representations and help in quick adaptation to different factors of non-stationarity across tasks. 

\textbf{Limitations: }In the current setting, our method relies on the slot attention mechanism to capture distinct objects in the environment. This implies that in the scenarios in which the slot attention mechanism is not able to bind to distinct objects in the pixel space (due to the size of objects, constant movement, etc.), our method OC-GVFs would not perform well without architectural or input representation tweaks. Slot attention is heavily dependent on separating objects by colors so in cases, where the colors of the objects are similar they would always be bound to the same slot which can be problematic depending on the task.

A promising direction for future work would be interesting to explore whether GVFs can perform zero-shot transfer only with the help of previously learned cumulants albeit in the presence of the environments with similar objects. In addition, instead of task-specific cumulants, cumulants can also be trained with task-agnostic losses, which might generalize even better across tasks.

\section*{Acknowledgements}
The authors would like to thank the ICML reviewers for valuable feedback on an earlier draft of the paper. In addition, we would like to express our sincere gratitude to Google, CIFAR (the Canadian Institute for Advanced Research), NSERC (the Natural Sciences and Engineering Research Council of Canada) and Canada Excellence Research Chairs (CERC) program for their invaluable support and funding. We are also immensely grateful to Compute Canada for providing the computational resources necessary to carry out the experiments. In addition, we would like to extend a special thank you to Alex Kearney for insightful discussions and feedback during the initial research phase.

\bibliographystyle{icml2023}
\bibliography{refs}

\newpage
\appendix
\section{Appendix}
\subsection{Domains}\label{sec:apndx_domains}
\begin{figure}[!ht]
    \centering
    \includegraphics[width=\linewidth]{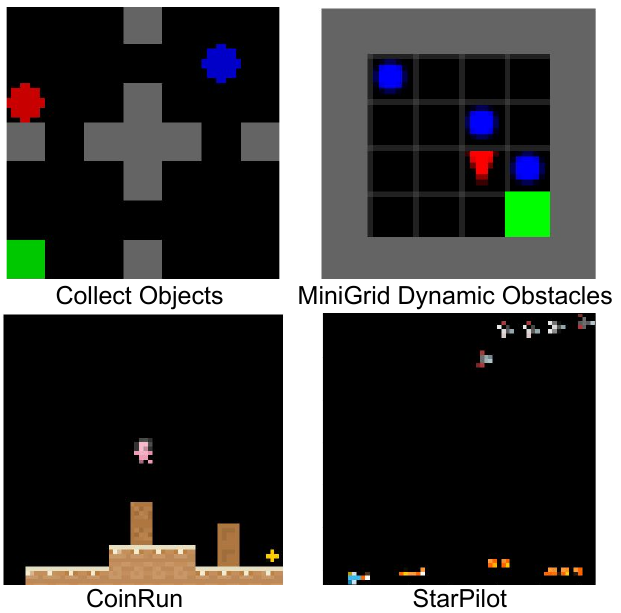}
    \caption{\textbf{Domains.} This figure illustrates the domains used in the Experiments. All of the environments require identifying objects in the input space from pixels.}
\label{fig:co_env}
\end{figure}

Fig.~\ref{fig:co_env} are sample input states of all the domains used in the experiments. 
\begin{figure*}[!ht]
\centering
\subfigure[\centering]{
    \centering
    \includegraphics[height=0.22\linewidth, width=0.5\linewidth]{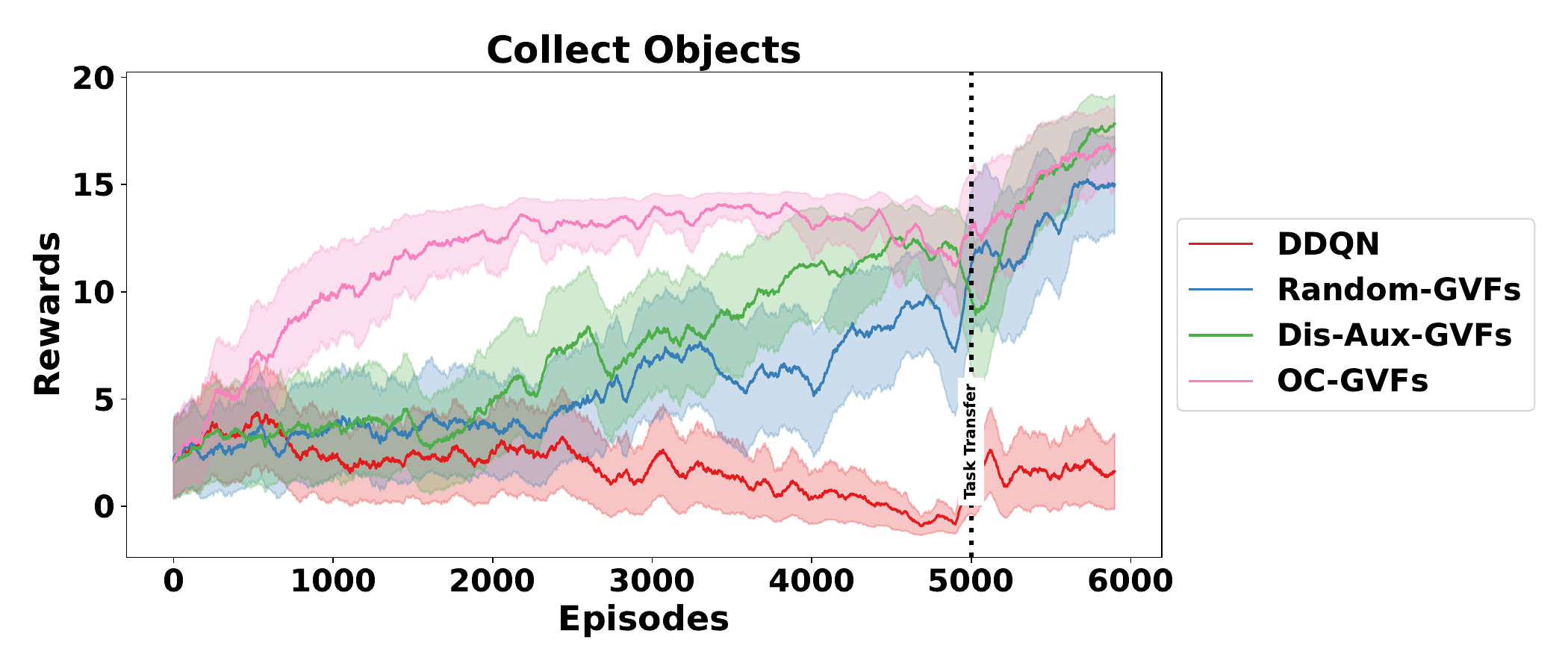}
    \label{fig:co_transfer}}
\qquad
\subfigure[\centering]{
    \centering
    \includegraphics[height=0.22\linewidth,width=0.43\linewidth]{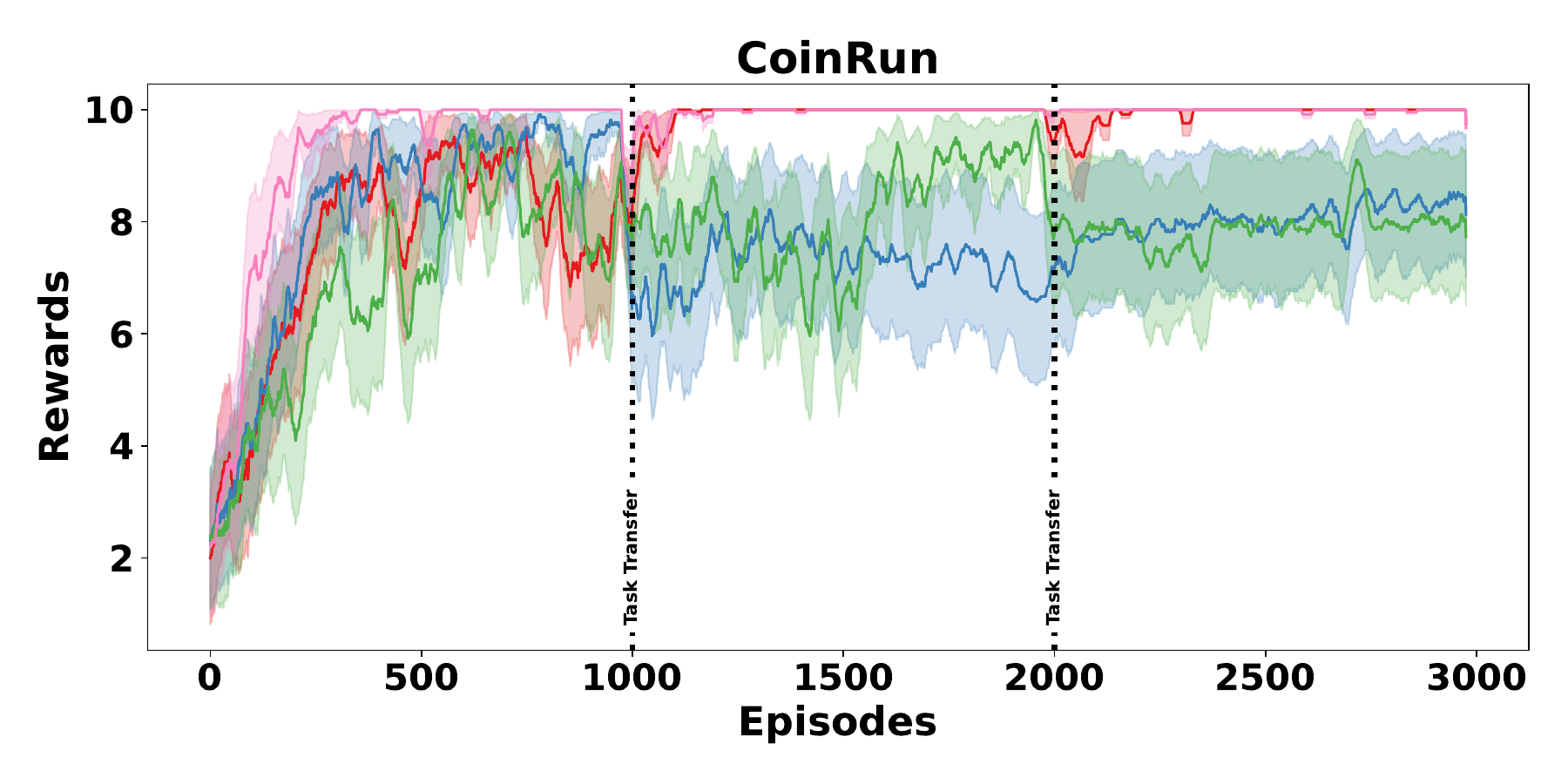}
    \label{fig:cr_transfer}}
    \vspace*{-5mm}

\caption{\textbf{Transfer Learning:} Our approach OC-GVFs can tackle the scenario of changing tasks better than the baselines with no appreciable drop in performance. In Collect Objects environment we change tasks by changing goal locations after certain episodes. Whereas, in Coinrun the different tasks refer to procedurally generated levels varying in task difficulty. Again we compare with the baselines mentioned above and report results over 10 seeds.}
\label{fig:transfer}
\end{figure*}


\subsection{Baseline Descriptions}\label{sec:apndx_baselines}
 Since we claim to learn object-centric representations in the pixel space, we use a 3 convolution layer architecture for the representation layers as in the original DQN implementation ~\cite{mnih2015humanlevel}. This is the part of the main network that was earlier explained in Sec.~\ref{sec:main} and the main RL network is a DQN agent learning just with the TD loss. In our setup, the question network is a simple MLP network with one fully-connected layer and it outputs the cumulants to be used in the main network as explained in the main paper. All the baselines used in the paper are explained in detail as follows.
\begin{enumerate}
    \item \textbf{DDQN}: For all our experiments, we used Double DQN~\cite{ddqn} as our base RL algorithm. This is the extension of the DQN~\cite{mnih2015humanlevel} algorithm with double Q learning to prevent maximization bias.
    \item \textbf{Random-GVFs}: We also compared performance against using random cumulants for the learned GVFs. For these experiments, we sampled cumulants from a uniform distribution between $[-1,1]$. The learned GVFs only help in learning the representations, and they are not used as main features.
    \item \textbf{HandCrafted-GVFs}: This is similar to Random GVFs except we use human knowledge to preselect good cumulants in advance. For the Collect Objects environments, the five cumulants chosen were +1 for reaching the red goal and each of the corridors. 
    \item \textbf{Discovery of Useful Questions as Auxiliary Tasks (Dis-Aux-GVFs)}: This algorithm is from~\citet{veeriah2019discovery} which discovers cumulants from the main RL loss via meta-gradient descent. This is the basic form of discovery that uses the main RL loss to learn cumulants given inputs.
    \item \textbf{GVFs as Features}: In all of these methods, the GVFs are not only used to influence the representation but are used as features of the main RL network. All these methods include concatenation in the latent space by a linear projection of the GVFs followed by layer normalization. In this category, we have four more algorithms:
    \begin{itemize}
        \item \textbf{Random GVFs as Features (Random-GVFs+)}
        \item \textbf{Hand Crafted GVFs as Features (HC-GVFs+)}
        \item \textbf{Using Discovered auxiliary tasks as features (Dis-Aux-GVFs+)}: This algorithm would be most similar to~\citet{metagvf} where the discovery architecture is kept similar to~\citet{veeriah2019discovery}, except we add compatibility with all forms of input space with the help of projected concatenation in the latent space and added stability with layer normalization.
        \item \textbf{Using discovered Object-Centric GVFs as Features (OC-GVFs)}: The proposed algorithm falls under the umbrella of using GVFs as features.
    \end{itemize}
\end{enumerate}
\textbf{Note}: Algorithms 2-4 all learn action value functions as GVFs similar to how it was designed in~\citet{veeriah2019discovery}. All algorithms falling under the umbrella of using GVFs as features (Alg 5) use state value functions as GVFs. Unstable or divergent action values can cause catastrophic failure particularly when they are used as features.

\subsection{Additional Experiments}
\subsubsection{Transfer Learning Experiments} \label{sec:apndx_transfer}
\begin{figure}[!ht]
    \centering    
    \subfigure[]{\includegraphics[height=0.3\linewidth, width=0.8\linewidth]{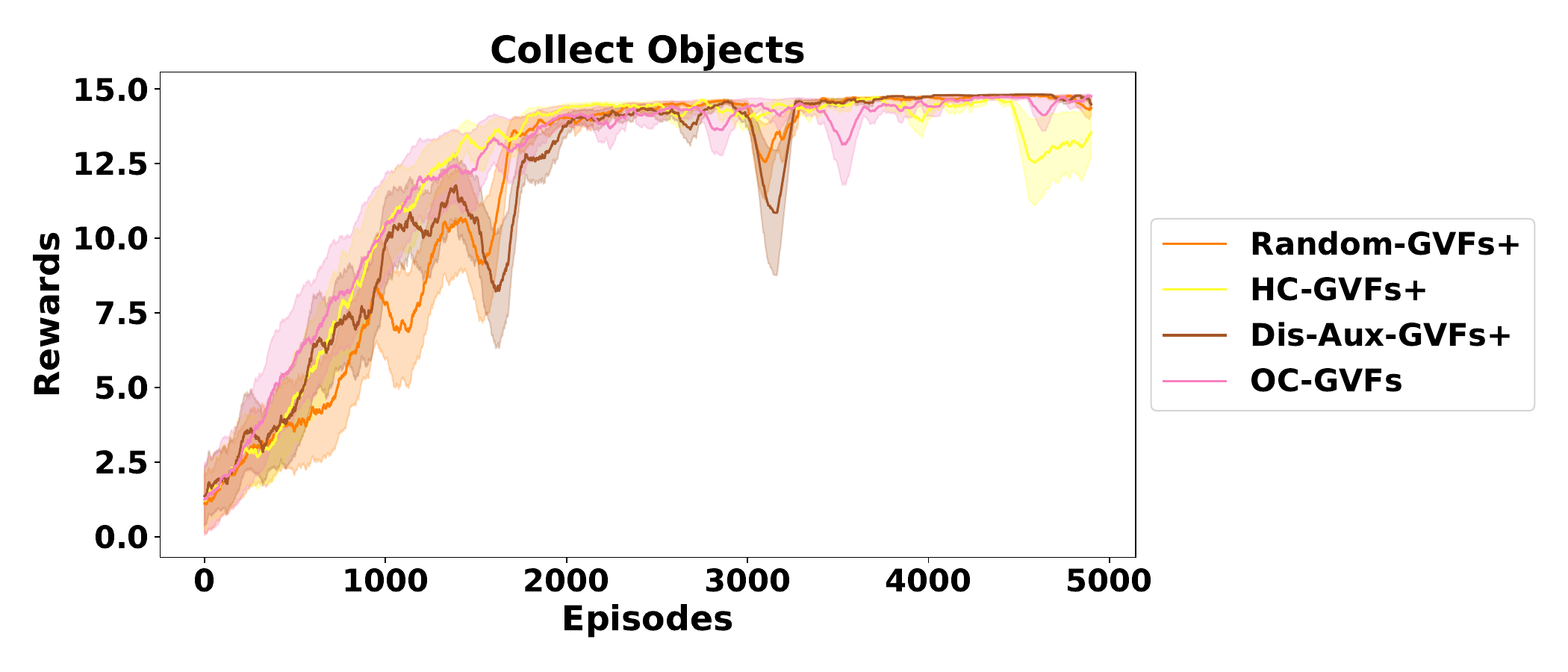}}
    \subfigure[]{\includegraphics[height=0.3\linewidth,width=0.8\linewidth]{media/CollectObjects/co_esp.pdf}}
    \subfigure[]{\includegraphics[height=0.3\linewidth,width=0.8\linewidth]{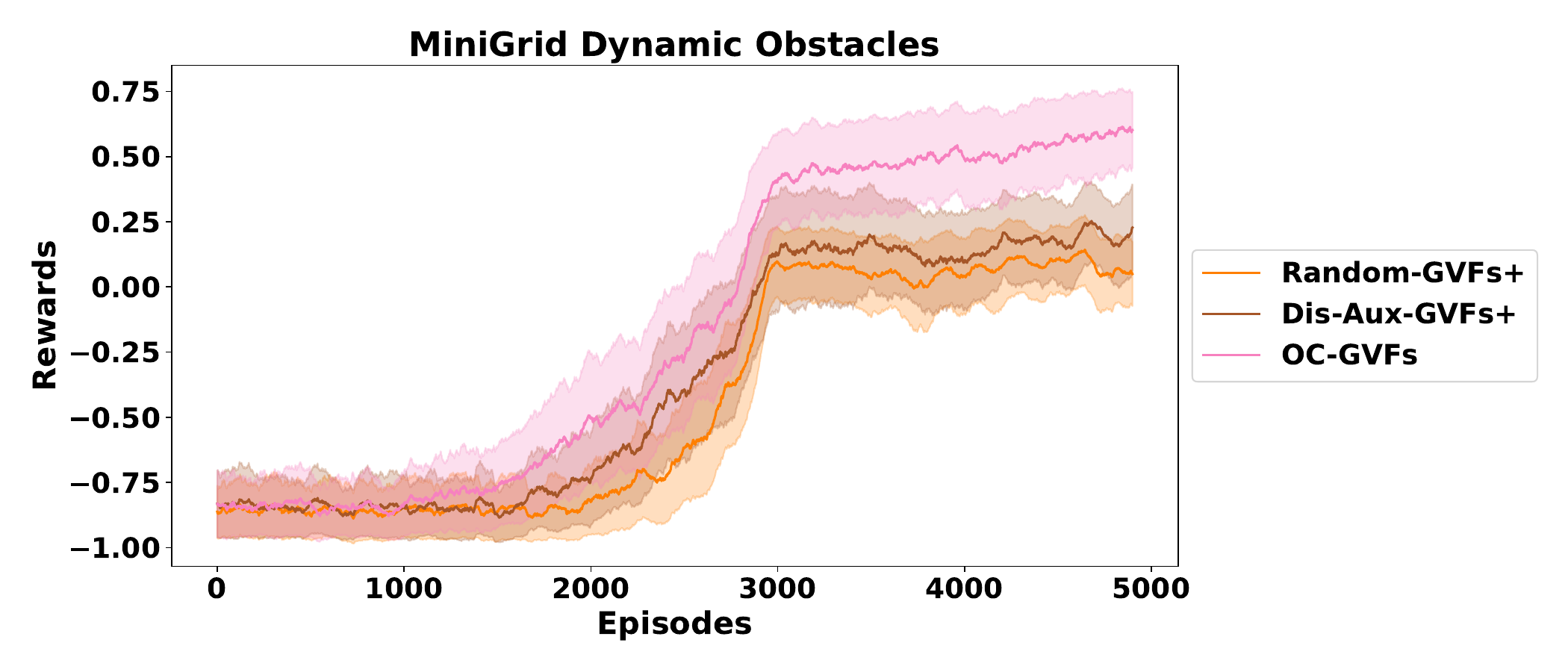}}
    \caption{Comparison of the baseline architectures on Collect Objects with stationary(top) \& non-stationary rewards (middle) using GVFs as features ($+$). Each of the baselines algorithms has \textbf{projected concatenation} and \textbf{layer normalization} added. In spite of these additions, the baselines do not perform well. Only our algorithm OC-GVFs is able to perform close to HandCrafted-GVFs with predefined knowledge of the environment. This further highlights the importance of object-centric discovery.}
    \label{fig:apndx_esp}
\end{figure}
In Fig.~\ref{fig:transfer}, we demonstrate  settings with increase complexity in tasks; 1) CollectObjects where the task changes once and and 2) CoinRun where the level of difficulty changes twice. In CollectObjects, the new task corresponds to adding one randomly positioned red goal, while CoinRun includes procedural generation of Level $0->2->3$. We see in Fig.~\ref{fig:transfer} that our approach OC-GVFs not only outperforms the other methods in the initial levels, but also is quick in few-shot adaptation when faced with new harder levels in comparison to the other baselines. DDQN performs poorly in the initial level, but once it is caught up, it remains consistent with the increasing complexity of levels, but suffers from a higher variance as compared to OC-GVFs. Other methods including \cite{veeriah2019discovery} struggle in all levels of CoinRun. In these settings, the baselines perform well because we believe this is a much simpler setting as even after transfer we allow the network to train for 1000 episodes which is enough to learn a good representation.


\subsubsection{Feature Discovery without object driven cumulant
learning} \label{sec:apndx_esp}
Fig.~\ref{fig:apndx_esp} shows the performance on Collect Objects with static and dynamic object locations and MiniGrid Dynamic Obstacles. In Collect Objects, all the algorithms do relatively well because it is an easy domain, however, we can see the utility of using object-centric representations as features in MiniGrid where the difference is more pronounced.
\subsubsection{Action versus State Values} \label{sec:apndx_vq}
\begin{figure}[!ht]
    \centering    \includegraphics[height=0.4\linewidth, width=0.8\linewidth]{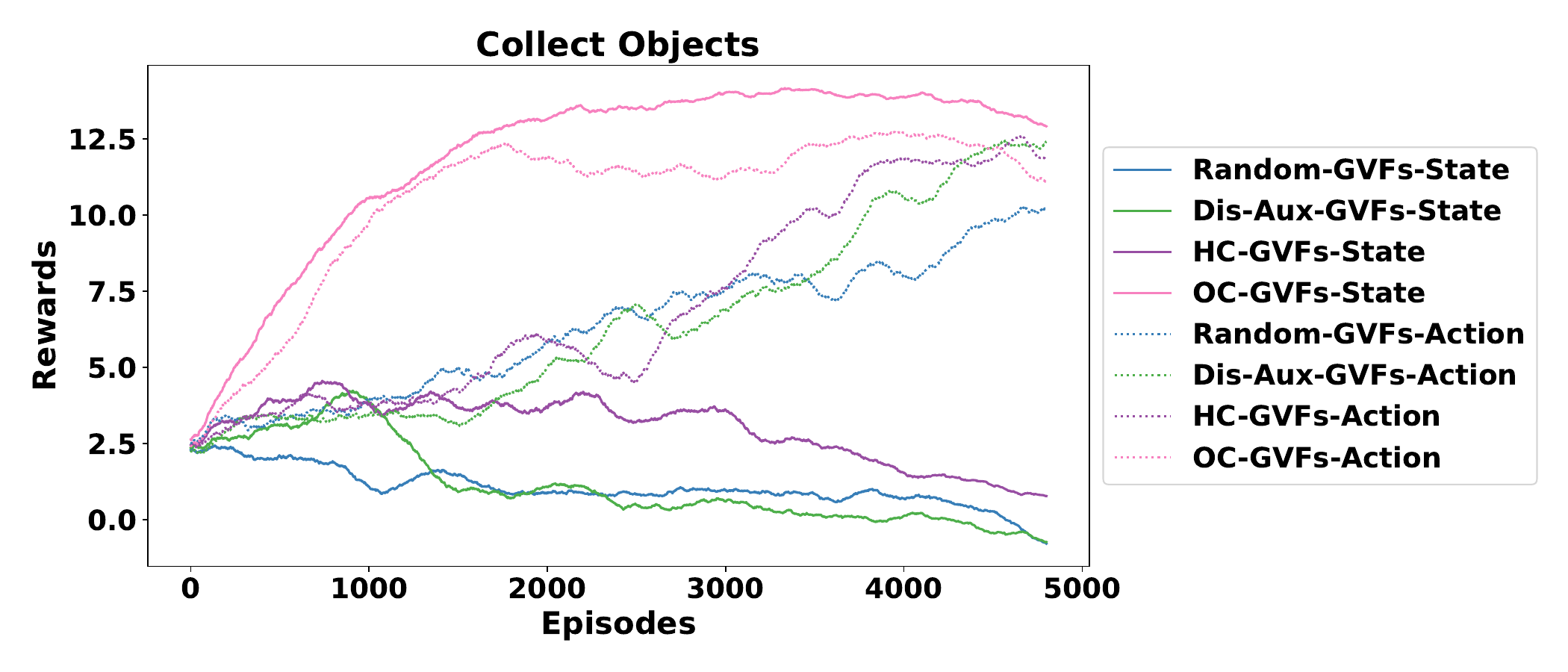}
    \caption{\textbf{State \& Action Values:} Comparison of baselines algorithms using state v/s action value GVFs. OC-GVFs are robust to both, however, with state values, the baseline algorithms are not as good due to the lack of information captured by them for learning a good representation.}
    \label{fig:vq}
\end{figure}

Fig.~\ref{fig:vq} demonstrates the performance of the baselines for both state and action value functions in the Collect Objects environment with random object placements.
As mentioned earlier in Sec.~\ref{sec:main}, GVFs can be learned with both state and action values. Action values are more prone to divergence because of the off-policy nature of action value functions and as a result using them as features results in a slightly worse performance for our algorithm. However, using state values did not seem to work at all for the other baselines when used only for training the common representation. We believe this is due to the less information provided by state value functions, which provide an expectation over all possible actions for each state.

\subsection{Implementation Details}\label{sec:apndx_implementation}
For the experiments, we optimized the performance of the main DDQN agent in terms of hyper-parameters and kept them constant for each of the baselines. The other parameters for discovery were kept the same as~\citet{veeriah2019discovery}. Most of the slot attention hyper-parameters were kept the same as~\citet{slotattn}. Some modifications were made in the encoder and decoders depending on the resolution of the input images. We used a much smaller resolution compared to~\citet{slotattn}, as finding separating them distinctly was not the objective for our method. This end-to-end pipeline will be undoubtedly more expensive because of the additional overhead of training slot attention. However, in our experiments, we resize the image to a sufficiently small (32x32) resolution which reduces training time and at the same time discovers slots that are `good enough' for learning cumulants. For CoinRun we used (64x64) images with lesser training steps. The final set of hyper-parameters are listed in Table~\ref{tab:hyps}. All our experiments were run on a single V100 GPU.

\begin{table*}[!ht]
\caption{Hyper-Parameters of all experiments}
\label{tab:hyps}
\centering
\begin{tabular}{|l|l|l|l|}
\hline
\textbf{Environment} &
  \textbf{Algorithm Parameters} &
  \textbf{Encoders and Decoders} &
  \textbf{Slot Attention Parameters} \\ \hline
  CollectObjects &
  \begin{tabular}[c]{@{}l@{}}``train\_episodes": 5000,\\ 
   ``batch\_size": 32,\\   
 ``target\_period": 100,\\   
 ``replay\_capacity": 100000,\\   
 ``hidden\_arch": {[}64,32{]},\\   
 ``epsilon\_begin": 1.0,\\   
 ``epsilon\_end": 0.01,\\   
 ``epsilon\_steps": 0.8,\\   
 ``discount\_factor": 0.99,\\   
 ``learning\_rate": 0.0001,\\   
 ``eval\_episodes": 100,\\  
 ``evaluate\_every": 50,\\  
 ``num\_gvfs": 5,\\  ``unroll\_steps": 10\end{tabular} &
  \begin{tabular}[c]{@{}l@{}}\textbf{Main CNN:}\\ Conv2D:filters=16, kernel=3\\ MaxPool2D:strides=2\\ Conv2D:filters=32, kernel=3\\ MaxPool2D:strides=2\\ Conv2D:filters=64, kernel=3\\ \textbf{Slot Attention Encoder:}\\ Conv2D\\ filters=32, kernel=3\\ filters=32, kernel=3\\ filters=64, kernel=3\\ \textbf{Slot Attention Decoder:} \\ Conv2DTranspose\\ filters=64, kernel=3, stride=2\\ filters=32, kernel=3, stride=2\\ filters=32, kernel=3, stride=1\\ filters=4, kernel=3, stride=1\end{tabular} &
  \begin{tabular}[c]{@{}l@{}}``sa\_batch\_size": 16,\\  
 ``sa\_resolution": 32,\\   
 ``sa\_num\_slots": 5,\\   
 ``sa\_num\_iterations": 3,\\  
 ``sa\_learning\_rate": 0.0004,\\   
 ``sa\_num\_train\_steps": 200000,\\  
 ``sa\_warmup\_steps": 10000,\\  
 ``sa\_decay\_rate": 0.5,\\  
 ``sa\_decay\_steps": 100000\end{tabular} \\ \hline
CoinRun &
  \begin{tabular}[c]{@{}l@{}}``train\_episodes": 5000,\\  ``batch\_size": 32,\\   ``target\_period": 100,\\   ``replay\_capacity": 10000,\\   ``hidden\_arch": {[}64,32{]},\\   ``epsilon\_begin": 1.0,\\   ``epsilon\_end": 0.01,\\   ``epsilon\_steps": 0.8,\\  ``discount\_factor": 0.99,\\   ``learning\_rate": 0.0001,\\   ``eval\_episodes": 100,\\   ``evaluate\_every": 50,\\   ``num\_gvfs": 5,\\   ``unroll\_steps": 10\end{tabular} &
  \begin{tabular}[c]{@{}l@{}}\textbf{Main CNN:}\\ Conv2D:filters=16, kernel=3\\ MaxPool2D:strides=2\\ Conv2D:filters=32, kernel=3\\ MaxPool2D:strides=2\\ Conv2D filters=64, kernel=3\\ \textbf{Slot Attention Encoder:}\\ Conv2D\\ filters=32, kernel=5\\ filters=32, kernel=5\\ filters=64, kernel=5\\ \textbf{Slot Attention Decoder:}\\ Conv2DTranspose\\ filters=64, kernel=5, stride=2\\ filters=32, kernel=5, stride=2\\ filters=32, kernel=5, stride=2\\ filters=32, kernel=3, stride=1\\ filters=4, kernel=3, stride=1\end{tabular} &
  \begin{tabular}[c]{@{}l@{}}``sa\_batch\_size": 16,\\   ``sa\_resolution": 64,\\   ``sa\_num\_slots": 5,\\   ``sa\_num\_iterations": 3,\\   ``sa\_learning\_rate": 0.0004,\\   ``sa\_num\_train\_steps": 100000,\\   ``sa\_warmup\_steps": 10000,\\   ``sa\_decay\_rate": 0.5,\\   ``sa\_decay\_steps": 100000\end{tabular} \\ \hline
\begin{tabular}[c]{@{}l@{}}MiniGrid\\ Dynamic\\ Obstacles\\\end{tabular} &
  \begin{tabular}[c]{@{}l@{}}``train\_episodes": 5000,\\   ``batch\_size": 32,\\   ``target\_period": 100,\\   ``replay\_capacity": 100000,\\   ``hidden\_arch": {[}64, 32{]},\\   ``epsilon\_begin": 1.0,\\   ``epsilon\_end": 0.001,\\   ``epsilon\_steps": 0.6,\\   ``discount\_factor": 0.99,\\   ``learning\_rate": 0.0001,\\   ``eval\_episodes": 100,\\   ``evaluate\_every": 50,\\   ``num\_gvfs": 5,\\   ``unroll\_steps": 10\end{tabular} &
  \begin{tabular}[c]{@{}l@{}}\textbf{Main CNN:}\\ Conv2D:filters=16, kernel=3\\ MaxPool2D:strides=2\\ Conv2D:filters=32, kernel=3\\ MaxPool2D:strides=2\\ Conv2D:filters=64, kernel=3\\ \textbf{Slot Attention Encoder:}\\ Conv2D\\ filters=32, kernel=3\\ filters=32, kernel=3\\ filters=64, kernel=3\\ \textbf{Slot Attention Decoder:}\\ Conv2DTranspose\\ filters=64, kernel=3, stride=2\\ filters=32, kernel=3, stride=2\\ filters=32, kernel=3, stride=1\\ filters=4, kernel=3, stride=1\end{tabular} &
  \begin{tabular}[c]{@{}l@{}}``sa\_batch\_size": 16,\\   ``sa\_resolution": 32,\\   ``sa\_num\_slots": 5,\\   ``sa\_num\_iterations": 3,\\   ``sa\_learning\_rate": 0.0004,\\   ``sa\_num\_train\_steps": 400000,\\   ``sa\_warmup\_steps": 10000,\\   ``sa\_decay\_rate": 0.5,\\   ``sa\_decay\_steps": 100000\end{tabular} \\ \hline
\end{tabular}
\end{table*}

\end{document}